\documentclass[runningheads]{llncs}

\let\llncssubparagraph\subparagraph
\let\subparagraph\paragraph
\usepackage[compact]{titlesec}
\let\subparagraph\llncssubparagraph

\usepackage{graphicx}
\usepackage{comment}
\usepackage{booktabs}
\usepackage{wrapfig, blindtext}
\usepackage{dsfont}
\usepackage{multirow}
\usepackage[T1]{fontenc}
\usepackage{pdflscape}
\usepackage{amsfonts,amsmath,amssymb}
\usepackage{float}
\usepackage{titlesec}
\usepackage{wrapfig}

\begin{document}
\title{Wearable-based Parkinson's Disease Severity Monitoring using Deep Learning}
\titlerunning{Wearable-based Parkinson's Disease Severity Monitoring}

\author{Jann Goschenhofer\inst{1, 2} \and Franz MJ Pfister \inst{1, 2} \and Kamer Ali Yuksel\inst{2} \and Bernd Bischl\inst{1}\and Urban Fietzek\inst{3,4} \and Janek Thomas\inst{1}}
\authorrunning{J. Goschenhofer et al.}

\institute{Dept. of Statistics, Ludwig-Maximilians University, Munich, Germany \\\email{jann.goschenhofer@campus.lmu.de} \and
ConnectedLife GmbH, Munich, Germany \and
Dept. of Neurology, Ludwig-Maximilians University, Munich, Germany \and
Dept. of Neurology and Clinical Neurophysiology, Schoen Clinic Schwabing, Munich, Germany}
\maketitle              
\begin{abstract}
One major challenge in the medication of Parkinson's disease is that the severity of the disease, reflected in the patients' motor state, cannot be measured using accessible biomarkers. Therefore, we develop and examine a variety of statistical models to detect the motor state of such patients based on sensor data from a wearable device. We find that deep learning models consistently outperform a classical machine learning model applied on hand-crafted features in this time series classification task. Furthermore, our results suggest that treating this problem as a regression instead of an ordinal regression or a classification task is most appropriate. For consistent model evaluation and training, we adopt the leave-one-subject-out validation scheme to the training of deep learning models. We also employ a class-weighting scheme to successfully mitigate the problem of high multi-class imbalances in this domain. In addition, we propose a customized performance measure that reflects the requirements of the involved medical staff on the model. To solve the problem of limited availability of high quality training data, we propose a transfer learning technique which helps to improve model performance substantially. Our results suggest that deep learning techniques offer a high potential to autonomously detect motor states of patients with Parkinson's disease.

\end{abstract}

\section{Introduction} 
\label{ch1}

Parkinson's disease (PD) is one of the most common diseases of the elderly and the second most common neurodegenerative disease in general after Alzheimer's \cite{SamaPerez-LopezRomagosaEtAl2012}. Two million Europeans are affected and 1\% of the population over the age of 60 in industrial nations are estimated to suffer from PD \cite{PringsheimJetteFrolkisEtAl2014,AhlrichsLawo2013}. Fortunately, the disease can be managed by applying the correct personalized dosage and schedule of medication, which has to be continuously adapted regarding the progress of this neurodegenerative disease. Crucial for the optimal medication is knowledge about the current latent motor state of the patients, which can not yet be measured effortlessly, autonomously and continuously. The motoric capabilities of the patients are distinguishable into three different motor states which can vary substantially over the course of a day within hours. The most prominent symptom is the tremor but the disease defining symptom ist the loss of amplitude and slowness of movement, also referred as bradykinesia \cite{PostumaBergSternEtAl2015}. In contrast to bradykinesia, an overpresence of dopaminergic medication can make affected patients execute involuntary excessive movement patterns which may remind an untrained observer of a bizarre dance. This hyperkinetic motor state is termed dyskinesia \cite{TsipourasTzallasFotiadisEtAl2011}. In a very basic approximation, people with Parkinson's disease (PwP) can be in three motor states: 1) the bradykinetic state (OFF), 2) a state without appearant symptoms (ON), and 3) the dyskinetic state (DYS) \cite{MarsdenParkes1976}. If the true motor state of PwP was known at all times, the medication dose could be optimized in such a way, that the patient has an improved chance to spend the entirety of his waking day in the ON state. An example for such a closed-loop approach can be found in Diabetes therapy, where the blood sugar level serves as a biomarker for the disease severity. Patients suffering from Diabetes can continuously measure their blood sugar level and apply the individual, correct medication dose of Insulin in order to balance the disease. Analogously, an inexpensive, autonomous and precise method to assess the motor state might allow for major improvements in personalized, individual medication of PwP.

Advancements in both wearable devices equipped with motion sensors and statistical modeling tools accelerated the scientific community in researching solutions for motor state detection of PwP since the early 2000s. In 1993, Ghika et al. did pioneering work in this field by proposing a first computer-based system for tremor measurement \cite{GhikaWiegnerFangEtAl1993}. A comprehensive overview on the use of machine learning and wearable devices in a variety of PD related problems was recently provided by Ahlrichs et al. \cite{AhlrichsLawo2013}. A variety of studies compare machine learning approaches applied on hand-crafted features with deep learning techniques where the latter show the strongest performance \cite{HssayeniBurackJimenez-ShahedEtAl2018,SamaPerez-LopezRomagosaEtAl2012,KeijsersHorstinkGielen2003,KeijsersHorstinkGielen2006,TsipourasTzallasFotiadisEtAl2011,HammerlaFisherAndrasEtAl2015,HssayeniBurackGhoraani2016,EskofierLeeDaneaultEtAl2016,UmPfisterPichlerEtAl2018}. In the present setting, a leave-one-subject-out (LOSO) validation is necessary to yield unbiased performance estimates of the models \cite{SaebLoniniJayaramanEtAl2017}. Thus, it is surprising that only a subset of the reviewed literature deploys a valid LOSO validation scheme \cite{HssayeniBurackJimenez-ShahedEtAl2018,HssayeniBurackGhoraani2016,UmPfisterPichlerEtAl2018,EskofierLeeDaneaultEtAl2016,TsipourasTzallasFotiadisEtAl2011}. It is noteworthy that one work proposes modeling approaches with a continuous response \cite{KeijsersHorstinkGielen2003}, while the rest of the literature tackles this problem as a classification task to distinguish between the different motor states. Amongst the deep learning approaches, it is surprising that none of the related investigations describe their method to tune the optimal amount of training epochs for the model, which is not a trivial problem as discussed in Section~$3.3$. A strutured overview on the related literature is given in Table \ref{tab:related_literature}.

\begin{table}[!bp]
\centering
\rotatebox{90}{%
\begin{minipage}{1.0\textheight}
\caption[Overview on results from related literature]{Overview on results from the literature on Motor State detection for PwP. In the method column, the MLP refers to a Multi-layer Perceptron, FE to manual feature extraction, SVM to a Support Vector Machine and LSTM for Long-short-term-memory network. In the label column, the names of the class labels are depicted. From this column, one can infer that only two authors used continuous labels and thus regression models for their task. Generally, a comparison of the reviewed approaches is difficult due to high variation in the data sets, methods and evaluation criteria.}
\resizebox{1.0\textwidth}{!}{%
    \begin{tabular}{@{}l l l l l l l l p{2.7cm} @{}}\toprule[2pt] Author & Method & Validation & Subjects & Sensors & Position & Setting & Labels & Results\\ \midrule[2pt]
    \cite{HssayeniBurackJimenez-ShahedEtAl2018} & FE, SVM & LOSO & 19 & 6 & wrist, ankle & lab & ON, OFF & Acc.: 90.5\%\\ \midrule[0.1pt]
    \cite{UmPfisterPichlerEtAl2018} & CNN & LOSO & 30 & 1 & wrist & free & OFF, ON, DYS & Acc.: 63.1\%\\ \midrule[0.1pt]
    \cite{SamaPerez-LopezRomagosaEtAl2012} & FE, SVM & Holdout Patients & 20 & 1 & belt & lab & ON, OFF & Acc.: 94.0\% \\ \midrule[0.1pt]
    \cite{HssayeniBurackGhoraani2016} & LSTM & LOSO & 12 & 1 & ankle & free & ON, OFF & Acc.: 73.9 \%\\ \cmidrule[0.1pt]{2-9}
     & FE, SVM & LOSO & 12 & 1 & ankle & free & ON, OFF & Acc.: 65.7 \%\\ \midrule[0.1pt]
    \cite{EskofierLeeDaneaultEtAl2016} & CNN & LOSO & 10 & 2 & wrist & lab & ON, OFF & Acc.: 90.9\% \\\midrule[0.1pt] 
    \cite{HammerlaHalloranPloetz2016} & FE, MLP & Leave-one-day-out & 34 & 2 & wrist & free & OFF, ON, DYS, Sleep & F1: 55\% \\ \midrule[0.1pt]
    \cite{HammerlaFisherAndrasEtAl2015} & FE, MLP & 7-fold CV & 34 & 2 & wrist & lab & OFF, ON, DYS, Sleep & F1: 76\% \\\midrule[0.1pt]
    \cite{KeijsersHorstinkGielen2006} & FE, MLP & Train set & 23 & 6 & trunk, wrist, leg & lab & ON, OFF & F1: 97\% \\\midrule[0.1pt]
    \cite{TsipourasTzallasFotiadisEtAl2011} & FE, MLP & LOSO & 29 & 6 & wrist, leg, chest, waist & lab & DYS Y/N & Acc.: 84.3\% \\\midrule[0.1pt]
    \cite{KeijsersHorstinkGielen2003} & FE, MLP & 80/20 Split & 13 & 6 & trunk, wrist, leg & free & Continuous & Acc.: 77\%\\ \midrule[0.1pt]
    \cite{LoniniDaiShawenEtAl2018} & FE, RF & LOSO & 20 & 1 & wrist & lab & ON, OFF & AUC: 0.73 \\\cmidrule[0.1pt]{2-9}
    & FE, RF & LOSO & 20 & 1 & wrist & lab & Tremor Y/N & AUC: 0.79 \\\midrule[0.1pt]
    Our approach \footnote{Performance measures are detailed in Section~$5$}& CNN & LOSO & 28 & 1 & wrist & free & Continuous & MAE: 0.77 \\ \cmidrule[0.1pt]{2-9}
    & CNN & LOSO & 28 & 1 & wrist & free & 9-class & $\pm1$ Acc.: 86.95\% \\
    \bottomrule[2pt]
\end{tabular}}
\label{tab:related_literature}
\end{minipage}}
\end{table}

\subsubsection{Contributions} This paper closes the main literature gaps in machine learning based monitoring of PD: the optimal problem setting for this task is discussed, a customized performance measure is introduced and a valid LOSO validation strategy is applied to compare time series classification (TSC) deep learning and classical machine learning approaches. Furthermore, the application of transfer learning strategies in this domain is investigated. 

This paper is structured as follows: The used data sets are described in Section~$2$. In Section~$3$, peculiarities of the problem are discussed. Furthermore, in Section~$4$ model architectures and problem settings are proposed and their results are discussed in Section~$5$. A transfer learning strategy is introduced in Section~$6$ to overcome the limited availability of training data. 

\section{Data} 
\label{ch2}

Data was collected from PwP to model the relation between raw movement sensor data and motor states. The acceleration and rotation of patient's wrists was measured via inertial measurement units (IMUs) integrated in the Microsoft band 2 fitness tracker \cite{Microsoft2018} with a standard frequency of 62.5Hz. The wrist was chosen as sensor location as it is the most comfortable location for a wearable device to be used in the patients' daily lifes and was shown to be sufficient for the detection of Parkinson-related symptoms \cite{LoniniDaiShawenEtAl2018,CurtzeNuttCarlson-KuhtaEtAl2015}. The raw sensor data was downsampled to a frequency of 20Hz as PD related patterns do not exceed this frequency \cite{HammerlaFisherAndrasEtAl2015}. A standard procedure in human activity recognition is the segmentation of continuous sensor data streams into smaller windows. As the data in this study was annotated by a medical doctor on a minute-level, the window length was set to one minute. To increase the amount of training data, the windows were segmented with an overlap of 80\% which is in line with related literature \cite{ZengNguyenYuEtAl2014,EskofierLeeDaneaultEtAl2016,HammerlaHalloranPloetz2016}. To neutralize any direction-specific information, the $L_2$-norms of the accelerometer and gyroscope measurements are used as model input, leading to two time series per window. Finally, the data was normalized to a $[0, 1]$ range via quantile transformation. 

We consider the machine learning problem of the feature space $\mathcal{X} \subset \mathbb{R}^p$, with $p = 1200\cdot2$, a target space $\mathcal{Y}$ described below and a performance measure $\mathcal{P}: \mathcal{Y} \times f(\mathcal{X}) \rightarrow \mathbb{R}$ measuring the prediction quality of a model $f: \mathcal{X} \rightarrow \mathcal{Y}$, trained on the data set $\mathcal{D} = \left \{ (x^{(1)}, y^{(1)}), ..., (x^{(n)}, y^{(n)})\right \}$ where a tuple $\left(x^{(i)}, y^{(i)}\right) \in \mathcal{X} \times \mathcal{Y}, i = 1, ..., n$ refers to a single labeled one minute window. 

The disease severity $\mathcal{Y}$ is measured on a combined version of the UPDRS \cite{GoetzTilleyShaftmanEtAl2008} and the mAIMS scale \cite{LaneGlazerHansenEtAl1985}. The UPDRS scale is based on a diagnostic questionnaire for physicians to rate the severity of the bradykinesia of PwP on a scale with $0$ representing the ON state to $4$, the severly bradykinetic state. The mAIMS scale is analogue to the UPDRS, but in contrast used for the clinical evaluation of dyskinetic symptoms. Both scales were combined and the UPDRS scale was flipped to cover the whole disease spectrum. The resulting label scale takes values in $\mathcal{Y} = \{-4, ..., 4\}$ where $y^{(i)}=-4$ means a patient is in a severely bradykinetic state, $y^{(i)}=0$ is assigned to a patient in the ON state and $y^{(i)}=4$ resembles a severely dyskinetic motor state. The sensor data was labeled by a medical doctor who shadowed the PwP during one day in a free living setting. Thus, the rater monitored each patient, equipped with an IMU, while they performed regular daily activities and the rater clinically evaluated the patients' motor state at each minute.

In total, 9356 windows were extracted from the data of $28$ PwP. By applying the above described preprocessing steps, the amount of windows was increased to $45944$.

\section{Challenges} 
\label{ch_challenges}

\subsection{Class imbalance}
\label{ch:class_imbalance}

\begin{wrapfigure}{r}{0.4\linewidth}
    \centering
    \vspace{-30pt}
    \includegraphics[width=0.4\textwidth]{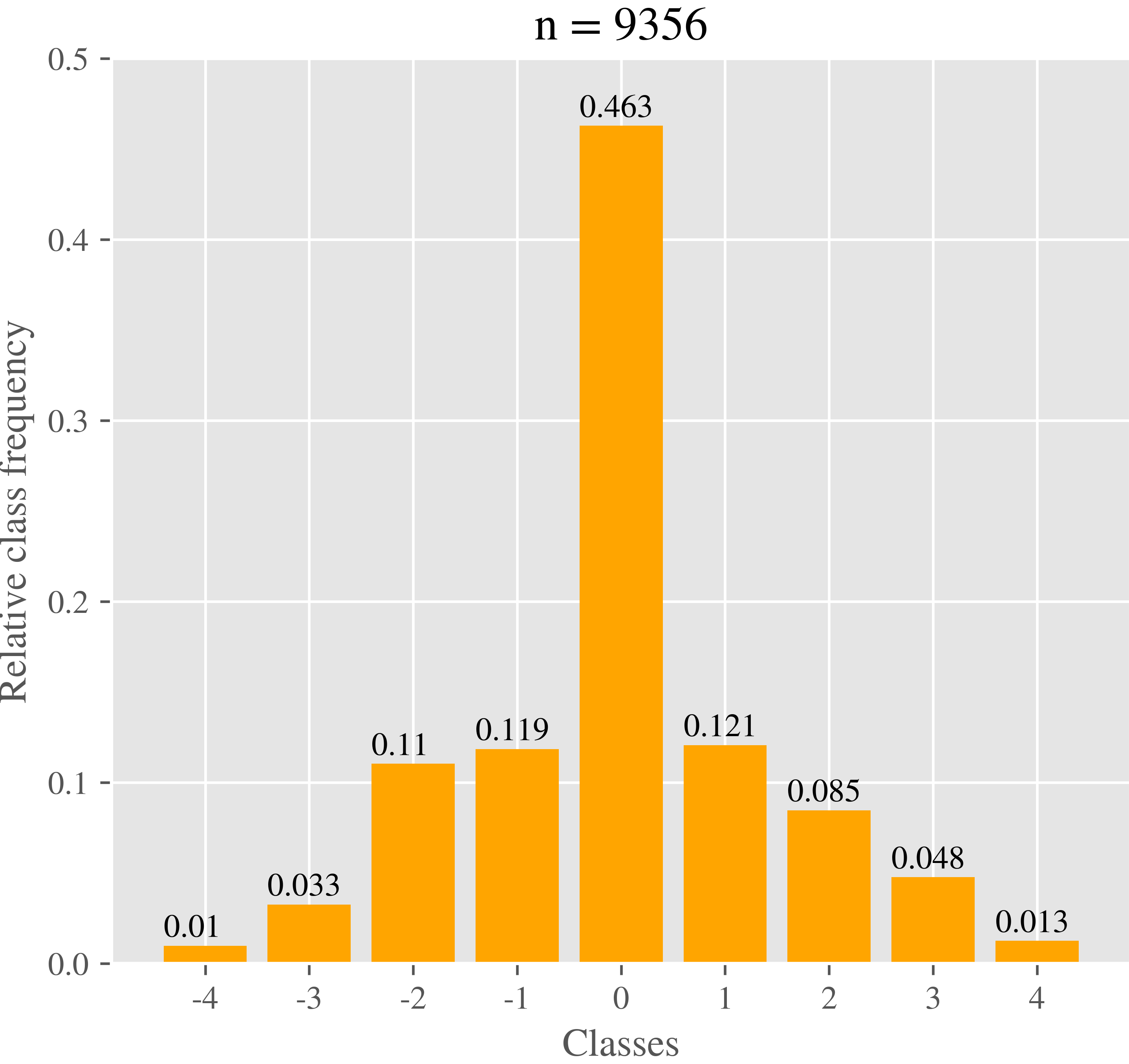}
    \caption[Class distribution]{Label distribution of the data which is highly centered around $y=0$.}
    \vspace{-15pt}
    \label{fig:class_imbalance}
\end{wrapfigure}

The labeled data set suffers from high label imbalance towards the center of the scale as shown in Figure~\ref{fig:class_imbalance}. Thus, machine learning models will be biased towards predicting the majority classes \cite{HeGarcia2009}. 

A straightforward way of dealing with this problem is to reweight the loss contribution of different training data samples. This way, the algorithm incurs heavier loss for errors on samples from minority classes than for those of majority classes, putting more focus on the minority classes during training. The weights for the classes $j \in \mathcal{Y} = \{-4, ..., 4\}$ are calculated as follows:

\begin{equation}
        c_j = \frac{n}{n_j}; \quad
        \tilde{c}_{j} = |\mathcal{Y}| \cdot \frac{c_j}{\sum_{j \in \mathcal{Y}}c_j}
\end{equation}

where $|\mathcal{Y}|$ describes the amount of classes, $n$ is the total amount of samples, $n_j$ the amount of samples for class $j$ and thus $c_j$ is the inverse relative frequency of class $j$ in the data. Further, the weights $c_j,j \in  \mathcal{Y}$ are normalized such that the sum of the weights is equal to the amount of classes. The individual weight of one sample is referred to as $\omega^{(i)}$ which is the normalized weight $\tilde{c}_{j}$ associated with the label $y^{(i)}$ of this sample $i$ such that $y^{(i)} = j$. 

\subsection{Custom Performance Measure}

It is crucial for the practical application of the final model to select an adequate performance measure which reflects the practical requirements on the model. Based on discussions with involved medical doctors, we found that larger errors should be penalized heavier which implies a quadratic error. Additionally, errors in the wrong direction of the scale, e.g. $\hat y^{(i)} = -1, y^{(i)} = 1$, should have a higher negative impact than errors with the same absolute distance in the correct direction, e.g. $\hat y^{(i)} = 3, y^{(i)} = 1$. The rationale behind this is that an exaggerated diagnostic evaluation which follows the true pathological scenario harms the patient less than an opposing one. Furthermore, the cost of predicting the wrong pathological direction increases with the severity of the disease: diagnostic errors weigh heavier on patients with strong symptoms compared to patients that are only mildly affected by the disease. In summary, three main requirements on the custom performance measure were identified: non-linearity, asymmetry and not being translation invariant.

Inspired by econometric forecasting \cite{ElliottTimmermannKomunjer2005}, the following asymmetric performance measure which satisfies the first two previous requirements is introduced:

\begin{equation}
    \label{final_loss}
    P_\alpha(\mathcal{D}, f) = \frac{1}{|\mathcal{D}|}\sum_{x^{(i)}, y^{(i)} \in \mathcal{D}}\left[\alpha + sign\left(y^{(i)} - f(x^{(i)})\right)\right]^2 \left( f(x^{(i)}) - y^{(i)}\right)^2
\end{equation}

where $\alpha \in [-1, 1]$ controls the asymmetry such that:

\begin{equation}
    \alpha
    \begin{cases}
        \in ]-1, 0], & \text{penalization of underestimation,}\\
        = 0, & \text{symmetric loss},\\
        \in [0, 1[, & \text{penalization of overestimation}.\\
    \end{cases}
\end{equation}

This performance measure is the squared error multiplied by a factor that depends on the parameter $\alpha$ and on the over- or underestimation of the true label via the $sign$ function. As motivated in the third requirement, the asymmetry should depend on the true label values. Therefore, $y$ is connected with $\alpha$ by introducing $\alpha^*$ such that $\alpha = \frac{y^{(i)}}{4} \alpha ^*$ where  $y^{(i)} \in \mathcal{Y} = \{-4, ..., 4\}$, hence $\alpha^* \in [0, 1]$. The constant denominator $4$ is used to link $\alpha$ and $\alpha^*$ in such a way that the sign of $\alpha$ that governs the direction of the asymmetric penalization is controlled by the true labels $y$. This leads to the formalization:

\begin{equation}
    \label{final_loss_2}
    P_{\alpha^*}(\mathcal{D}, \hat f) = \frac{1}{|\mathcal{D}|}\sum_{x^{(i)}, y^{(i)} \in \mathcal{D}} \left[\frac{y^{(i)}}{4} \alpha ^* + sign\left(y^{(i)} - \hat f(x^{(i)})\right)\right]^2 \left(\hat f(x^{(i)}) - y^{(i)}\right)^2
\end{equation}

\begin{figure}
\centering
\begin{minipage}[t]{.50\textwidth}
\centering
\includegraphics[width=1.0\textwidth,height=4.5cm]{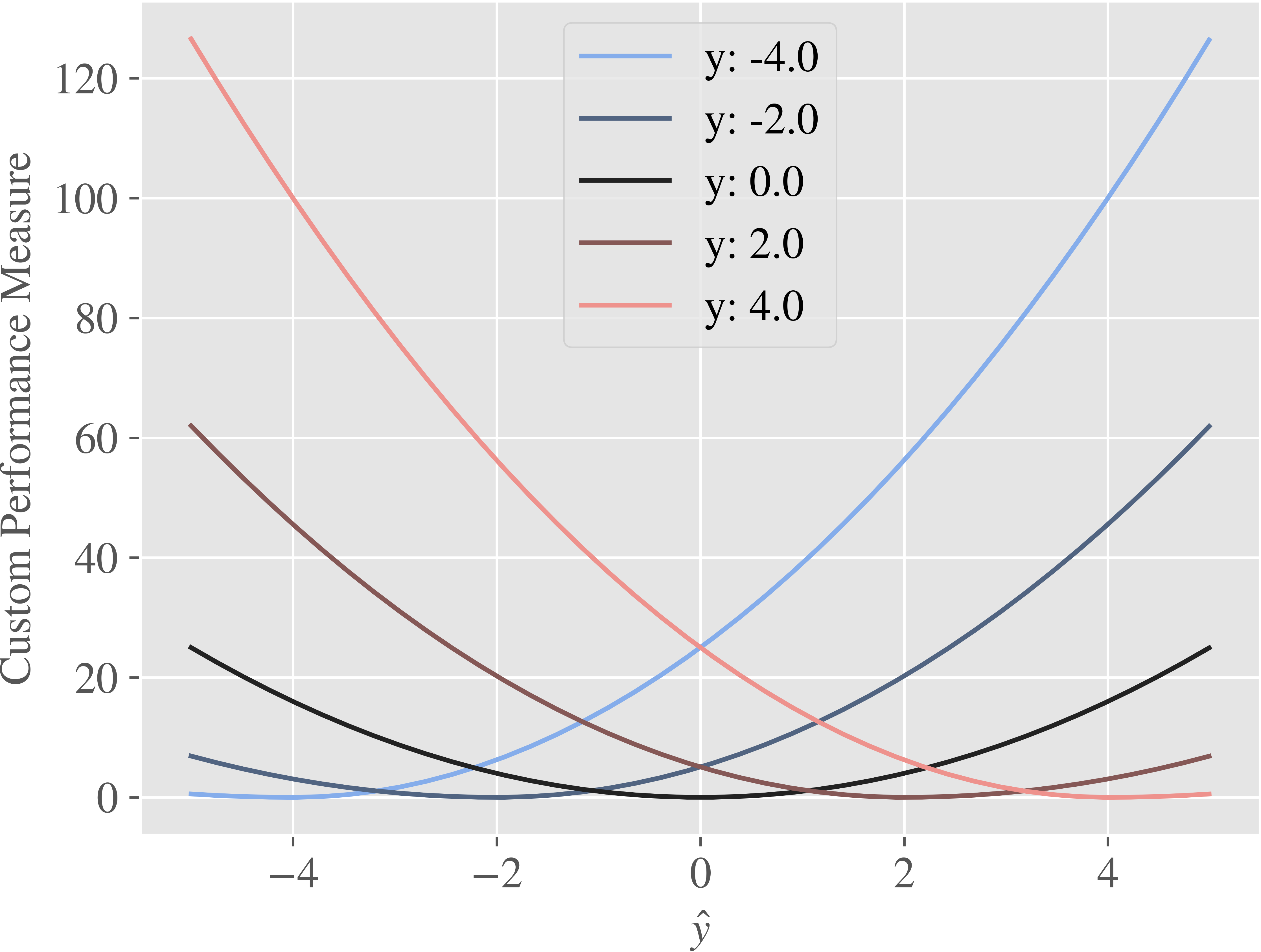}
\caption{Behavior of the performance measure $P_{\alpha^* = 0.25}$ on the y-axis for different labels $y$ and the corresponding predictions $\hat y$ on the x-axis.}
\label{fig:signum_behavior}
\end{minipage}\hfill
\begin{minipage}[t]{.45\textwidth}
\centering
\includegraphics[width=.8\textwidth,height=4.5cm]{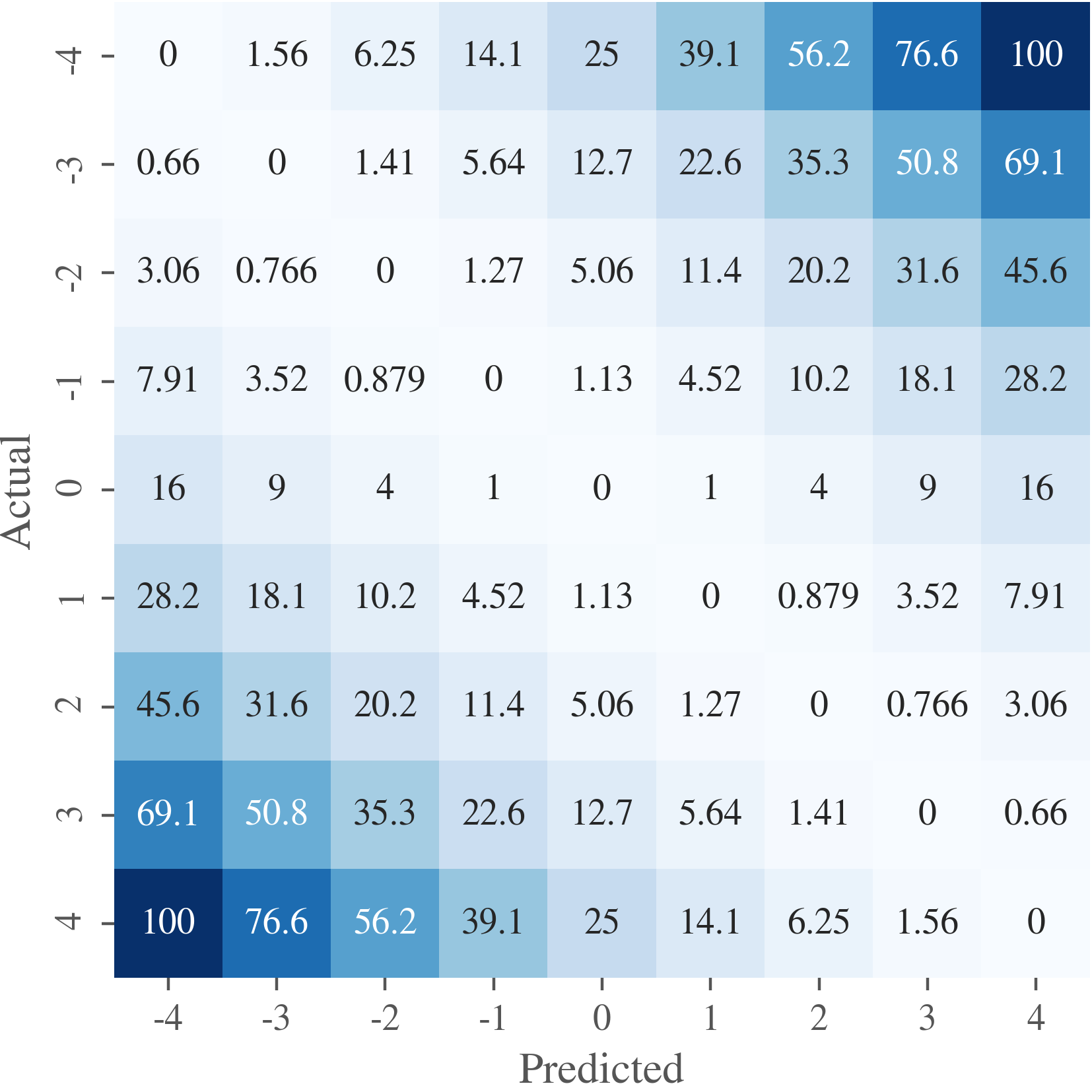}
\caption{Cost factors resulting from $P_{\alpha^* = 0.25}$ that are associated with each combination of actual and predicted values.}
\label{fig:costmatrix}
\end{minipage}
\end{figure}

The parameter $\alpha^* = 0.25$ was set based on the feedback of the involved medical experts\footnote{Feedback was collected by comparing multiple cost matrices as shown in Figure~\ref{fig:costmatrix}.}. The model will be heavily penalized for the overestimation of negative labels and for the underestimation of positive labels. For instance, the performance measure for $y^{(i)}=2$ and prediction $\hat y^{(i)}= 1$ is higher (1.265) than for $\hat y^{(i)} = 3$ (0.765). The asymmetry of the measure is reciprocally connected to the magnitude of the label $y$ in both, the negative as well as the positive direction, e.g. for $y^{(i)} = 1$ it is more symmetric than for $y^{(i)} = 3$. Furthermore, $P_{\alpha^*}$ collapses to a regular quadratic error for $y^{(i)}=0$. The behavior of the measure is further illustrated in Figure~\ref{fig:signum_behavior}.

\subsection{Leave-one-subject-out validation}

Proposed models are expected to perform well on data from patients not seen before. Using regular cross validation (CV) strategies, subject-specific information could be exploited resulting in an overly optimistic estimate of the generalization performance \cite{SaebLoniniJayaramanEtAl2017}. Consequently, a standard leave-one-subject-out (LOSO) validation scheme is often applied in settings were much data are gathered from few subjects \cite{BaoIntille2004,FisherHammerlaPloetzEtAl2016,EskofierLeeDaneaultEtAl2016}. Thereby, a model is trained on all except one subject and then tested on the left out subject, yielding an unbiased performance estimate. This is repeated for each individual subject and all resulting estimates are averaged.

The usage of early stopping \cite{GoodfellowBengioCourville2016} requires the introduction of a tuning step to determine the optimal amount of training epochs $e^*$ in each of the LOSO folds, which in turn requires a second inner split of the data set. In a setting with unlimited computational resources, one would run a proper LOSO validation in the inner folds, determine $e^*$, train the model on the whole data except the left out subject and evaluate the trained model on that subject. With a total amount of 28 patients, this would result in the training of $28\cdot27 = 756$ models for the validation of one specific architecture. As a cheaper solution, the first 80\% one minute windows per patient are used for training and the last 20\% for early stopping. 

\section{Problem Setting, Models, Class Imbalance} 
\label{ch4}

\subsection{Problem Setting}

As explained in Section \ref{ch2}, the target was measured on a discrete scale $y \in \mathcal{Y} = \{-4, ..., 4\}$ where $y = -4$ represents severe bradykinesia, $y = 0$ the ON state and $y = 4$ severe dyskinesia. This gives rise to the question whether the problem should be modeled as a classification, an ordinal regression or a regression task. The majority of previous research in this domain treats the problem as binary sub-problems with the goal to just detect whether the PwP experience symptoms, regardless of their severity. The granular labeling scheme used here follows an ordinal structure. For instance, a patient with $y=-4$ suffers from more severe bradykinesia than one with $y=-3$. In contrast, simple multi-class classification treats all class labels as if they were unordered. A simple way of including this ordinal information is to treat the labels as if they were on a metric scale and apply standard regression methods. However, this implies a linear relationship between the levels of the labels. For example, a change in the motor state from $y=-4$ to $y=-3$, $\delta_{-4, -3}$, could have a different meaning than $\delta_{-2, -1}$, though they would be equivalent on a metric scale. The formally correct framing of such problems is ordinal regression which takes into account the ordered structure of the target but does not make the strong linearity assumption \cite{GutierrezPerez-OrtizSanchez-MonederoEtAl2016}. This model class is methodologically located at the intersection of classification and metric regression. All three problem settings are compared in Section~$5$.\\

\subsection{Models}

\textbf{Random Forest} A Random Forest \cite{Breiman2001} was trained on manually extracted features from the raw sensor data, similar to related literature \cite{HammerlaFisherAndrasEtAl2015,EskofierLeeDaneaultEtAl2016,HssayeniBurackGhoraani2016,SamaPerez-LopezRomagosaEtAl2012}. From each sample window of both signal norms, a total of $34$ features such as mean, variance and energy were extracted (a complete list can be found in the digital Appendix). This is a standard procedure in TSC  \cite{ChristKempa-LiehrFeindt2016,CasalePujolRadeva2011}. The Random Forest was specifically chosen as a machine learning baseline due to its low dependency on hyperparameter settings and its strong performance in general.

\textbf{FCN} The Fully Convolutional Net (FCN) was introduced as a strong baseline deep learning architecture for TSC \cite{WangYanOates2017}. The implementation resembles that of Wang~et~al. except that the final layer consists of $|\mathcal{Y}|=9$ or $1$ neuron(s) for classification and regression, respectively.  

\textbf{FCN Inception} Inception modules led to substantial performance increases in computer vision and are motivated by the observation that the kernel size of the convolutional layers are often chosen rather arbitrarily by the deep learning practitioner \cite{SzegedyLiuJiaEtAl2015}. The rationale is to give the model the opportunity to choose from different kernel sizes for each convolutional block and distribute the amount of propagated information amongst the different kernels. One inception module consists of branches with with kernel sizes $1,5,7$ and $13$ respectively and a depth of $64$ each, plus one additional max-pooling branch with a kernel size of $3$, followed by a convolution block with depth $64$ and a kernel size $1$. The final FCN Inception architecture essentially follows the original FCN with simple convolutional layers being replaced by 1D inception modules.   

\textbf{FCN ResNet} Similar to the inception modules, the introduction of residual learning has met with great enthusiasm in the deep learning community \cite{HeZhangRenEtAl2015}. The main advantage of such Residual Networks (ResNet) over regular CNNs is the usage of skip-connections between subsequent layers. These allow the information to flow around layers and skip them in case they do not contribute to the model performance, which makes it possible to train much deeper networks. Unlike inception modules, this model class was already adapted for TSC and proven to be a strong competitor for the original FCN \cite{WangYanOates2017}. The FCN ResNet was shown to outperform the standard FCN especially in multivariate TSC problems \cite{FawazForestierWeberEtAl2018}. Others argue that the ResNet is prone to overfitting and thus found it to perform worse than the FCN \cite{WangYanOates2017}. For the comparison in Section~$5$, three residual modules are stacked where each of the modules is identical to the standard FCN in order to provide comparability among architectures. The module depths were chosen as proposed by Wang~et~al. \cite{WangYanOates2017}.  

\textbf{FCN Broad} Pathologically, the disease severity changes rather slowly over time. Thus, it can be hypothesized that additional input information and a broader view on the data could be beneficial for the model. This model is referred to as FCN Broad and includes the following extension: the raw input data from the previous sample window $x_{t-1}$ and the following sample window $x_{t+1}$ are padded to the initial sample window $x_t$, which results in a channel depth of $6$ for the input layer.   

\textbf{FCN Multioutput} A broad variety of techniques for ordered regression exist \cite{HerbrichGraepelObermayer1999,FrankHall2001,ChenZhangDongEtAl2017,NiuZhouWangEtAl2016}. As a neural network based approach for ordered regression is required, a simple architecture is to create a single CNN, which is trained jointly on a variety of binary ranking-based sub-tasks \cite{NiuZhouWangEtAl2016}. A key element to allow the network to exploit the ordinal structure in the data is a rank-based transformation of labels. The categorical labels $y \in \mathcal{Y}$ are transformed into $K = |\mathcal{Y}|-1$ rank-based labels by:

\begin{equation}
    \label{eq:rank_label}
    y_{k}^{(i)} = 
    \begin{cases}
        1, \text{ if } y^{(i)} > r_k \\
        0, \text{ otherwise, }\\
    \end{cases}
\end{equation}

where $r_k$ is the rank for the $k$-th sub-problem for $k\in \{1, ..., K\}$. Following this label transformation, a multi-output CNN architecture is proposed where each of the $K$ outputs refers to one binary ranking-based sub-task. These are optimized jointly on a single CNN corpus. Thus, the sub-task $k$ is trained on a binary classification problem minimizing the binary cross entropy loss. The total model output consists of $K$ probability outputs for each input sample. In order to train the CNN jointly on those sub-tasks, the individual losses are combined to one cumulative loss: 

\begin{equation}
    \label{eq:multitask_XE}
    \begin{split}
    L^{\text{ranks}}(y^{(i)}, f(x^{(i)})) &= \sum_{k = 1}^{K} L_{k}^{b}(y_k^{(i)}, \hat y_k^{(i)}) \\
    \end{split}
\end{equation}

where $L_{k}^b$ is the binary cross-entropy loss for sub-task output $\hat y_k^{(i)}$. For inference, the $K$ outputs are summed up such that $\hat y^{(i)} = \sum_{k = 1}^{K} \hat y_k^{(i)} - 4$, where the scalar $4$ is subtracted from the sum over all probability outputs to map the predictions back to the initial label scale, yielding a continuous output.   

\textbf{FCN Ordinal} A second ordinal regression model can be created by training a regular FCN with an additional distance-based weighting factor in the multi-class cross entropy loss $L^{\text{m}}$:

\begin{equation}
L^{\text{ordinal}}(y^{(i)}, f(x^{(i)})) = \left|y^{(i)} - \hat y^{(i)}\right| \cdot L^{\text{m}}(y^{(i)}, \hat y^{(i)} )
\end{equation}

This way, the model is forced to learn the inherent ordinal structure of the data as it is penalized higher for predictions that are very distant to the true labels.

\section{Results} 
\label{ch5}

The models described in Section~$4$ were implemented in pytorch \cite{PaszkeGrossChintalaEtAl2017}. Model weights were initialized by Xavier-uniform initialization \cite{GlorotBengio2010} and ADAM \cite{KingmaBa2014} (learning rate = 0.00005, $\beta_1=0.9$, $\beta_2=0.99$) was used for training with a weight decay of $10^{-6}$. The performances of the models were compared in a LOSO evaluation as discussed in Section~$3.3$, using the performance measure $P_{\alpha^*=0.25}$ as introduced in Section~$3.2$. Finally, the sequence of motor state predictions is smoothed via a Gaussian filter whose $\mu$ and $\sigma$ parameters were optimized using the same LOSO scheme that was used for model training. The results are summarized in Table~\ref{tab:model_comparison}. An additional majority voting model which constantly predicts $\hat y=0$ is added as a naive baseline. 

\begin{table}[!h]
    \centering
    \caption[Results for model comparison]{Results for different models in multiple problem settings, measured using the performance measure introduced in Section~$3.2$ evaluated by LOSO validation. Additional commonly used performance measures are shown for completeness where the MAE is reported in a class-weighted (w. MAE) and a regular version and $\pm1$ Acc. refers to accuracy relaxed by 1 class.}
    \begin{tabular}{@{}llrrrrrr@{}}\toprule[2pt]
    Frame & Model & $P_{\alpha^*=0.25}$ & F1 & Acc. & $\pm1$ Acc. & w. MAE & MAE \\  \cmidrule[0.1pt]{1-3} \cmidrule[0.1pt](l){4-8}

    \multirow{1}{*}{\parbox{2.5cm}{Baseline}} 
    & Majority vote & 2.900 & 0.293  &0.702&0.463&0.661&0.960\\ \cmidrule[0.1pt]{1-3} \cmidrule[0.1pt](l){4-8}

    \multirow{2}{*}{\parbox{2.5cm}{Classification}} 
    & FCN & 0.800   &0.366 &0.809&0.340&0.312&0.890\\ \cmidrule[0.1pt]{2-3} \cmidrule[0.1pt](l){4-8}
    & Random Forest & 1.542    &0.394&0.802&0.459&0.465&0.802\\ \cmidrule[0.1pt]{1-3} \cmidrule[0.1pt](l){4-8}

    \multirow{2}{*}{\parbox{2.5cm}{Ordinal}} 
    & FCN & 0.752 &0.321&0.767&0.302&0.311&0.985\\ \cmidrule[0.1pt]{2-3} \cmidrule[0.1pt](l){4-8}
    & Multioutput FCN & 0.922 &0.361&0.820&0.352&0.344&0.873\\ \cmidrule[0.1pt]{1-3} \cmidrule[0.1pt](l){4-8}
    
    \multirow{2}{*}{\parbox{2.5cm}{Regression}} 
    & FCN  & \textbf{0.635} &0.346&0.843&0.338&0.293&0.836\\ \cmidrule[0.1pt]{2-3} \cmidrule[0.1pt](l){4-8}
    & FCN Inception  & 0.726 &0.380&0.841&0.370&0.304&0.842 \\ \cmidrule[0.1pt]{2-3} \cmidrule[0.1pt](l){4-8}
    & FCN ResNet & 0.841 &0.334&0.809&0.309&0.336&0.924\\ \cmidrule[0.1pt]{2-3} \cmidrule[0.1pt](l){4-8}
    & FCN Broad & 0.673 &0.347&0.835&0.339&0.294&0.852\\ \cmidrule[0.1pt]{2-3} \cmidrule[0.1pt](l){4-8}
    & Random Forest & 1.310 &0.411&0.848&0.436&0.423&0.760\\
    \bottomrule[2pt]
    \end{tabular}
    \label{tab:model_comparison}
\end{table}

The FCN was applied in all three problem settings. From Table~\ref{tab:model_comparison}, one can observe that regression performs better than ordered regression and classification. Similar results were obtained for the Random Forest baseline, where regression is superior to classification. It seems that the simple assumption of linearity between labels does not have a derogatory effect and a simpler model architecture as well as training process is of larger importance.

The comparison of the deep learning models with the Random Forest offers another interesting finding. For both, regression and classification, all deep learning models outperform the classic machine learning models. This finding justifies the focus on deep learning approaches and is in line with previous research discussed in the Introduction.

Niu~et~al. \cite{NiuZhouWangEtAl2016} claim that the Multioutput CNN architecture outperforms regular regression models in ordinal regression tasks. This can not be supported by the current results as the Multioutput FCN shows weaker performance than each of the deep learning architectures in the regression frame.

Looking at the results from the regression setting, one can observe that the simple FCN manages to outperform all more complex architectures as well as the Random Forest baseline. This could be explained by the increased complexity of these models: the FCN consists of $283,145$ weights, while the FCN Inception contains $514,809$ and the FCN ResNet $512,385$ weights. This problem is aggravated by the limited amount of training data.

\section{Transfer Learning} 
\label{ch_transfer_learning}

One of the most important requirements for the successful training of deep neural networks with strong generalization performance is the availability of a large amount of train data. Next to strong regularization and data set augmentation, one prominent method to fight overfitting and improve the model's generalization performance is transfer learning \cite{YosinskiCluneBengioEtAl2014}. A model architecture is first trained on source task $\mathcal{D}_A$. The learned knowledge, manifested in the model's weights, is used to initialize a model that should be trained on the target task $\mathcal{D}_B$. The model is then fine-tuned on $\mathcal{D}_B$ which often leads to faster model convergence and, dependent on the similarity of the tasks to an improvement in model performance. Though TSC is still an emerging topic in the deep learning community, first investigations into the adoption of transfer learning to time series data have been made \cite{FawazForestierWeberEtAl2018a}.

As a source task for the motor state detection, we train the model to classify between one-minute windows that were either gathered from PwP or from healthy patients. Therefore, we use a weakly labeled data set that contains $70175$ one-minute windows of sensor data along with the binary target if the corresponding patient suffers from Parkinson's disease or not. Among those patients, $50\%$ were healthy and $50\%$ suffered from PD. All proposed deep learning models were trained on this task and their weights were used for initialization. The final training on the actual data was done in the exact same fashion as described in Section~$5$.  

As shown in Table~\ref{tab:transfer_learning}, the transfer learning approach consistently improved the performance of all tested FCN architectures. This strategy also helped to further push the best achieved performance by the regression FCN. Thus, the pretrained FCN model in the regression setting is the overall best performing model.

\begin{table}[!h]
    \centering
    \caption[Performance of the transfer learning approaches]{Performance of the transfer learning approaches compared to their non-pretrained counterparts. Transfer learning consistently improves model performances. Additional commonly used measures are shown for the pretrained models only where the MAE is reported in a class-weighted (MAE w.) and a regular version and $\pm1$ Acc. refers to accuracy relaxed by 1 class.}
    \begin{tabular}{@{}llrrr rrrrr@{}}\toprule[2pt]
    Frame & Model & \multicolumn{2}{c}{$P_{\alpha^*=0.25}$} & Gain & F1 & Acc. & Acc. & MAE & MAE\\  
    &&regular&transfer& &&&$\pm 1$&w.&\\    \cmidrule[0.1pt]{1-5} \cmidrule[0.1pt](l){6-10}
    
    Classification & FCN & 0.800& 0.771 & 0.029 &0.375&0.361&0.813&0.318&0.897\\  \cmidrule[0.1pt]{1-5} \cmidrule[0.1pt](l){6-10}
    
    \multirow{2}{*}{Ordinal} 
    & FCN & 0.752 & 0.616 & 0.136 &0.350&0.326&0.802&0.295&0.921\\ \cmidrule[0.1pt]{2-5} \cmidrule[0.1pt](l){6-10}
    & Multioutput FCN & 0.922 & 0.657 & 0.265 &0.367&0.360&0.829&0.301&0.857\\ \cmidrule[0.1pt]{1-5} \cmidrule[0.1pt](l){6-10}

    Regression & FCN & 0.635 & \textbf{0.600} & 0.035 &0.407&0.388&0.870&0.273&0.772\\ 
    \bottomrule[2pt]
    \end{tabular}
    \label{tab:transfer_learning}
\end{table}

Transfer learning has the biggest effect on the performance of the Multioutput FCN, which indicates that this model requires a higher amount of training data. This is reasonable as it is arguably the most complex model considered. Further increasing the amount of training data might improve these complex models even more. 

\begin{figure}[!h]
    \centering
    \includegraphics[width=0.5\textwidth]{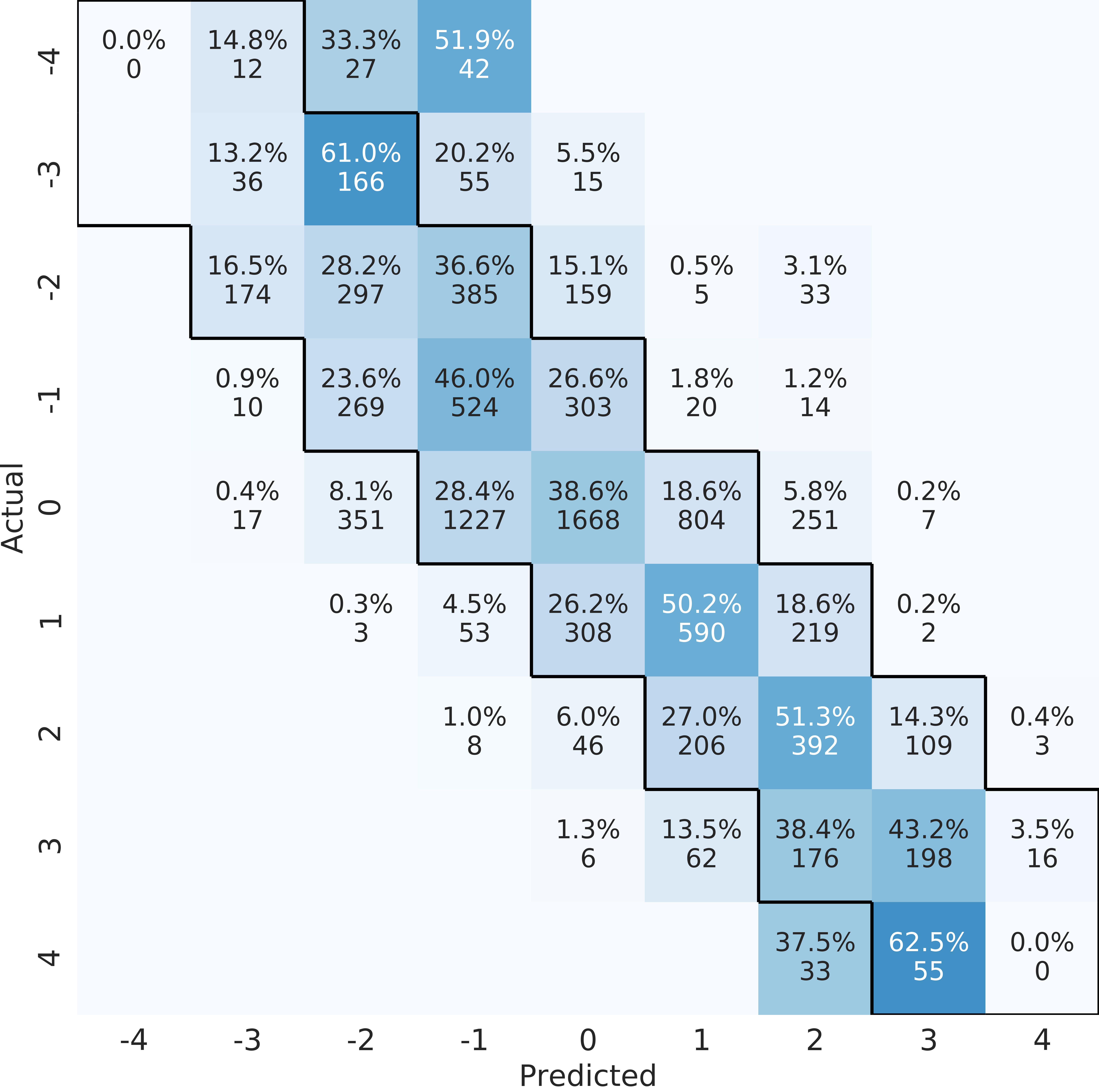}
    \caption[Confusion matrix]{Row-normalized confusion matrix for predictions from the pretrained regression FCN. Predicted continuous scores were rounded to integers. Allowing for deviations of $\pm1$ (framed diagonal region) yields a relaxed accuracy of 86.96\%.}
    \label{fig:confusion}
\end{figure}

\begin{figure}[!h]
    \centering
    \includegraphics[width=1\textwidth]{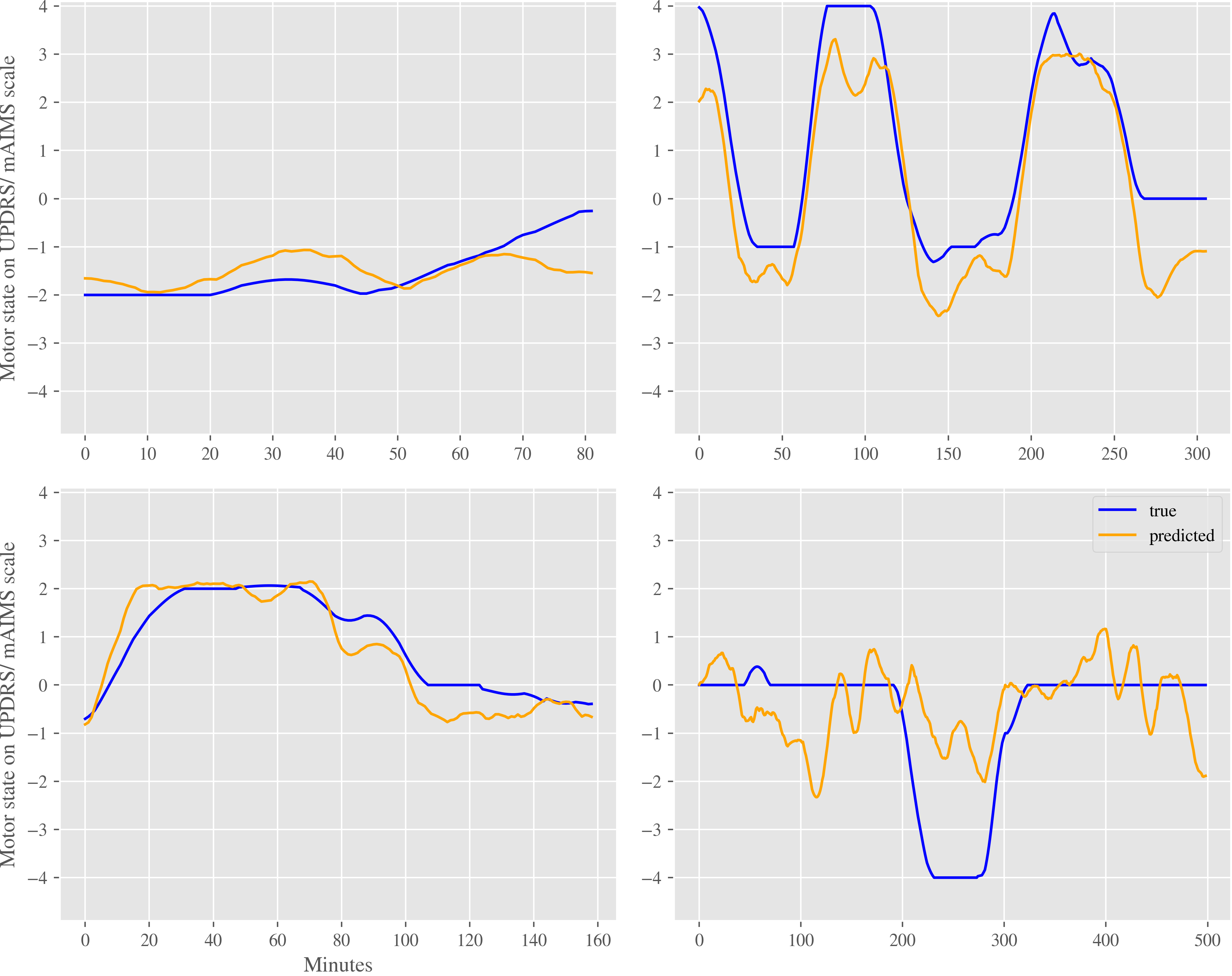}
    \caption[Day Curves]{Comparison of true (blue) and  predicted (orange) motor state sequences of four exemplary patients. The label scores are depicted on the y-axis and the minutes on the x-axis. The final model is able to capture the intra-day motor state regime changes of the PwP as shown on the top right plot. Still, the model fails to correctly detect the motor states in some patients e.g. the bottom right one.}
    \label{fig:curves}
\end{figure}

Some resulting predictions from the best performing model are illustrated in Figure \ref{fig:curves} and a confusion matrix of the model predictions is shown in Figure \ref{fig:confusion}. It is noteworthy that despite the class weighting scheme and the transfer learning efforts, the final model fails in correctly predicting the most extreme class labels. 

\section{Conclusion} 
\label{ch_conclusion}

Different machine learning and deep learning approaches were evaluated on the task to detect motor states of PwP based on wearable sensor data.
While the majority of related literature handles the problem as a classification task, the high quality and resolution of the provided data allows evaluation in different problem settings. Framing the problem as a regression task was shown to result in better better performance than ordered regression and classification. Evaluation was done using a leave-one-patient-out validation strategy on $28$ PwP using a customized performance measure, developed in cooperation with medical experts in the PD domain. The deep learning approaches outperformed the classic machine learning approach. Furthermore, the comparatively simple FCN offered the most promising results. A possible explanation would be that these intricate models call for more available data for successful training. Since high quality labeled data are scarce and costly in the medical domain, this is not easily achievable. First investigations into transfer learning approaches were successfully employed and showed model improvements for the deep learning approaches.

There exist a plethora of future work to investigate. Computational limitations made it impossible to evaluate all possible models in all problem settings as well as investigate recurrent neural network approaches. The successful usage of a weakly labeled data set for transfer learning suggests further research on the application of semi-supervised learning strategies. This work clearly shows the difficulty in fairly and accurately comparing existing approaches, as available data, problem setting and evaluation criteria differ widely between publications. The introduced performance measure could be a step into the right direction and can hopefully become a reasonable standard for the comparison of such models. In future work, one could directly use this performance measure as a loss function to train deep neural networks instead of using it for evaluation only.

\subsubsection{Acknowledgements} This work was financially supported by ConnectedLife GmbH and we thank the Schoen Klinik Muenchen Schwabing for the invaluable access to medical expert knowledge and the collection of the data set. This work has been partially supported by the German Federal Ministry of Education and Research (BMBF) under Grant No. 01IS18036A, and by an unrestricted grant from the Deutsche Parkinson Vereinigung (DPV) and the Deutsche Stiftung Neurologie.
\bibliographystyle{splncs04}

\end{document}